\pdfoutput=1

\documentclass[11pt]{article}

\usepackage[dvipsnames]{xcolor}
\usepackage{acl}
\usepackage{times}
\usepackage{latexsym}
\usepackage{}
\usepackage[utf8]{inputenc}
\usepackage[T1]{fontenc}
\usepackage{hyperref}
\usepackage{url}
\usepackage{booktabs}
\usepackage{amsfonts}
\usepackage{nicefrac}
\usepackage{microtype}
\usepackage{multirow}
\usepackage{algorithm}
\usepackage{algpseudocode}
\usepackage{amsmath}
\usepackage{amsthm}
\usepackage{enumitem}
\usepackage{hhline}
\usepackage{pdfpages}
\usepackage{pifont}
\usepackage{setspace}
\usepackage{soul}
\usepackage{tabularx}
\usepackage{graphicx}

\usepackage{amsmath,amsfonts,bm}

\def\eqref#1{equation~\ref{#1}}

\def\1{\bm{1}}

\DeclareMathAlphabet{\mathsfit}{\encodingdefault}{\sfdefault}{m}{sl}
\SetMathAlphabet{\mathsfit}{bold}{\encodingdefault}{\sfdefault}{bx}{n}

\title{\Large \methodname: Nested Human-in-the-Loop Reward Learning for\\Few-shot Language Model Control}

\author{%
    \textbf{Xiang Fan$^1$\thanks{\hphantom{*}Correspondence: \texttt{xiangfan00@gmail.com}} \hspace{.5em} Yiwei Lyu$^2$ \hspace{.3em} Paul Pu Liang$^1$} \\
    \textbf{Ruslan Salakhutdinov$^1$ \hspace{.3em} Louis-Philippe Morency$^1$}\vspace{0.1em}\\
    $^1$Carnegie Mellon University  \hspace{.5em} $^2$University of Michigan
}

\newcommand{\methodname}{\textsc{Nano}}
\begin{document}

\maketitle

\begin{abstract}
Pretrained\hspace{-.3pt} language\hspace{-.3pt} models\hspace{-.3pt} have\hspace{-.3pt} demonstrated extraordinary capabilities in language generation. However, real-world tasks often require controlling the distribution of generated text in order to mitigate bias, promote fairness, and achieve personalization. Existing techniques for controlling the distribution of generated text only work with quantified distributions, which require pre-defined categories, proportions of the distribution, or an existing corpus following the desired distributions. However, many important distributions, such as personal preferences, are unquantified. In this work, we tackle the problem of generating text following arbitrary distributions (quantified and unquantified) by proposing \methodname, a few-shot human-in-the-loop training algorithm that continuously learns from human feedback. \methodname~achieves state-of-the-art results on single topic/attribute\hspace{-.1pt} as\hspace{-.1pt} well\hspace{-.1pt} as\hspace{-.1pt} quantified\hspace{-.1pt} distribution control compared to previous works. We also show that \methodname~is able to learn unquantified distributions, achieves personalization, and captures differences between different individuals' personal preferences with high sample efficiency. Our code is available at \url{https://github.com/sfanxiang/Nano}.

\end{abstract}

\vspace{-3mm}
\section{Introduction}
\vspace{-1mm}

\begin{figure*}[t]
\vspace{0mm}
\includegraphics[width=\textwidth]{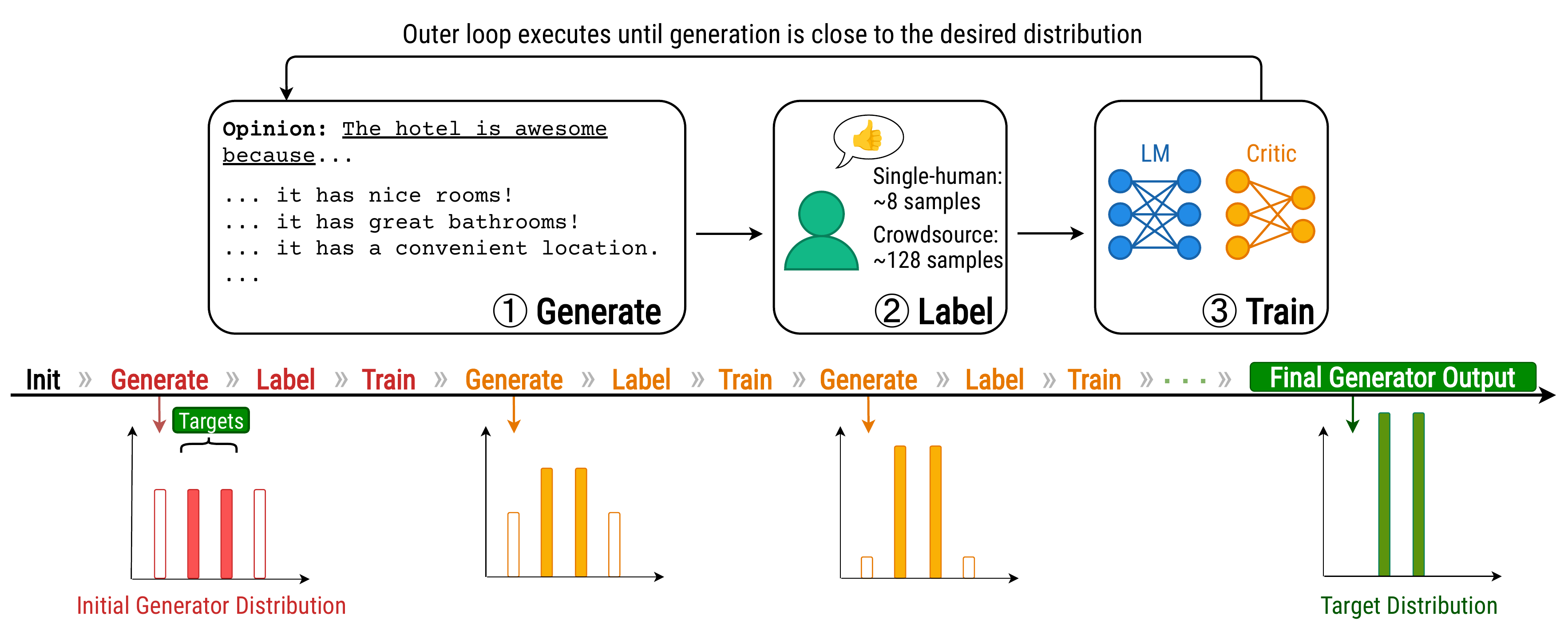}
\vspace{-6mm}
\caption{Overview of \methodname, a controllable text generation algorithm with two nested loops. The outer loop of our algorithm is a cycle of three learning phases: (1) generation, (2) human feedback, and (3) training. These allow the generation quality to improve over time.}
\vspace{-3mm}
\label{fig:overview}
\end{figure*}

Recent developments in large language models~\cite{radford2019language,gpt3} have advanced the state of automated text generation. However, to apply them to real-world tasks, it has become increasingly desirable to reduce social bias exhibited in large language models~\cite{Bender2021Dangers}, improve fairness~\cite{baldini-etal-2022-fairness}, and fit to diverse individual preferences~\cite{xue2009user}. These desired properties are only defined over a set of generated text instead of individual sentences. Therefore, they require control over the distribution of generated text~\cite{CNTRL_NLG_ICLR2021}. Existing works on distribution control deal with \textbf{quantified distributions}: they require knowledge of a known number of categories associated with each data point, an existing corpus following the desired distribution~\cite{gaopersonalized,Wang_2018,li-tuzhilin-2019-towards}, or a well-defined distribution with known proportions~\cite{CNTRL_NLG_ICLR2021} (such as $x\%$ category A, $y\%$ category B, etc.). However, \textbf{unquantified distributions}, such as arbitrary subjective distributions (e.g. ``news I find surprising'' for an arbitrary person), are relatively understudied. Because many distributions, including personal preferences, are fundamentally unquantified \textit{a priori}, the ability to learn unquantified distributions in a few-shot manner is key to modeling these distributions.

\begin{table}
    \footnotesize
    \begin{tabularx}{\linewidth}{>{\hsize=.3\hsize}X X}
        \hline
        \textbf{Our} \newline \textbf{personalized} \newline \textbf{generation} & \texttt{\underline{The hotel is very} \underline{awesome because} it is located in a \textbf{great neighborhood} accessible to the rest of the city.} \\
        \hline
        GPT-2\ \cite{radford2019language} & \texttt{\underline{The hotel is very} \underline{awesome because} I always feel like I can get a better experience.} \\
        \hline
    \end{tabularx}
    \vspace{-2mm}
    \caption{Comparison between personalized generation from \methodname\ vs. GPT-2. Our model is able to capture personal preferences with few-shot learning.}
    \label{tab:personalized_comp}
    \vspace{-4mm}
\end{table}

Our key insight for tackling arbitrary distributions is to continuously learn from intermediate human feedback, which points us towards the right direction at every step, instead of learning the final categories in one step. To this end, we propose \textbf{\underline{N}ested Hum\underline{an}-in-the-L\underline{o}op Reward Learning} (\textbf{\methodname}), a few-shot controllable text generation algorithm with two nested loops: the outer loop is a cycle of three learning phases (generation, human feedback, and training), and we introduce an inner loop in the generation phase, where we perform a tree search with nodes sampled from a language model, to address the issue of lack of samples. Furthermore, we find that human-in-the-loop training not only enables learning unquantified distributions, but also improves performance on quantified distributions. Our contribution is summarized as follows:
\vspace{-6.1mm}
\begin{itemize}[leftmargin=*]
    \item We introduce a \textbf{human-in-the-loop reward learning algorithm} that learns to generate text following arbitrary distribution through human feedback. We demonstrate that our method works for all of the following types of distributions: \textbf{single-topic/attribute}, \textbf{quantified distributions}, and \textbf{unquantified distributions}.
    \item We show that \methodname~is able to learn unquantified distributions, successfully achieves personalization, and captures differences between different individuals' personal preferences with only 64 labels from each person (RQ1).
    \item We achieve state-of-the-art result on controlling quantified distributions (RQ2) as well as single topic/attribute generation (RQ3) compared to previous works, while using only few-shot samples.
    \item Through ablation studies, we demonstrate the necessity of multi-iteration human feedback for high sample efficiency (RQ4) and justify our architecture's design choices (RQ5). We also show that our method extends to newer and larger language models than GPT-2.
\end{itemize}
\vspace{-1mm}
An illustration of our method is shown in Figure~\ref{fig:overview}, and a comparison of \methodname's capabilities to previous works is provided in Table~\ref{tab:comparison_related_work}.

\vspace{-1mm}
\section{Related Work}
\vspace{-2mm}

\begin{table*}[t]
\fontsize{8}{11}\selectfont
\setlength\tabcolsep{1.0pt}
  \begin{center}
    \renewcommand\tabularxcolumn[1]{m{#1}}
    \newcolumntype{H}{>{\setbox0=\hbox\bgroup}c<{\egroup}@{}}
    \newcommand\tablecmark{\centering\color{JungleGreen}\ding{51}}
    \newcommand\tablexmark{\centering\color{BrickRed}\ding{55}}
    \newcommand\tableqmark{\centering\color{RedOrange}\textbf{\texttt{?}}}
    \begin{tabularx}{1.0\textwidth}{X X X X X X X H}
      \hline
      & \centering \methodname~& \centering PPLM \cite{Dathathri2020Plug} & \centering GDC \cite{CNTRL_NLG_ICLR2021} & \centering Ziegler \cite{Ziegler2019FineTuningLM} & \centering InstructGPT \cite{instructgpt} & \centering QUARK \cite{lu2022quark} & \\
      \hline
      \baselineskip=1em No Reliance on \newline External Model & \tablecmark & \tablecmark & \tablecmark & \tablecmark & \tablecmark & \tablexmark & \\
      \hline
      \baselineskip=1em Topic/Sentiment \newline Control (RQ3) & \tablecmark & \tablecmark & \tablecmark & \tablecmark & \tablecmark & \tablecmark & \\
      \hline
      \baselineskip=1em Quantified \newline Distribution \newline Control (RQ2) & \tablecmark & \tablexmark & \tablecmark & \tablexmark & \tablexmark & \tablexmark & \\
      \hline
      \baselineskip=1em Unquantified \newline Distribution \newline Control (RQ1) & \tablecmark & \tablexmark & \tablexmark & \tablexmark & \tablexmark & \tablexmark & \\
      \hline
      \baselineskip=1em Fewshot \newline Learning (RQ4) & \tablecmark & \tablexmark & \tablexmark & \tablexmark & \tablecmark & \tablexmark & \\
      \hline
      \baselineskip=1em Personalization (RQ1) & \tablecmark & \tablexmark & \tablexmark & \tablecmark & \tablexmark & \tablexmark & \\
      \hline
    \end{tabularx}
   
  \end{center}
   \vspace{-4.5mm}
  \caption{Comparison of \methodname\ with related work. \methodname\ is able to work with arbitrary target distribution of text, regardless of domain, quantifiability, with no need for existing dataset following the distribution or external classifiers, and only requires human annotators to annotate a limited amount of sentences.}
  
  \label{tab:comparison_related_work}
   \vspace{-3.5mm}
\end{table*}

\textbf{Text generation models} are models designed to generate natural language. Natural language generation tasks include prompt completion, text summarization, translation, style transfer, etc. Current state-of-the-art language models include large transformer-based models, such as GPT-2~\cite{radford2019language} and GPT-3~\cite{gpt3}. These models are pre-trained on large corpus of text with masked token prediction, and can be easily fine-tuned to perform various text generation tasks as well as classification tasks. GPT-neo~\cite{gao2020pile} is one version of GPT that is specifically designed to allow few-shot learning of tasks. Recent advancements in text generation models also allows text generation models to follow text instructions, such as InstructGPT~\cite{instructgpt}. Before transformer-based models, natural language generation via template-based methods or hand-coed grammar-based systems~\cite{gatt2018survey} has also been explored. In our paper, we use GPT-2 ($355$M) as our baseline model.

\textbf{Controllable text generation} are techniques to generate text in a controllable fashion. Previous works have aimed to control generation towards specific topics or attributes (including classifier-based approach~\cite{Dathathri2020Plug} and reinforcement learning based approach~\cite{lu2022quark}) and control style of generated text via style transfer (including statistical NLP methods~\cite{hovy1987generating,xu2012paraphrasing}, neural generative models~\cite{prabhumoye2018style,lample2018multipleattribute,he2020probabilistic}, Retrieve-and-Edit approaches~\citep{li2018delete,hashimoto2018retrieve,guu2018generating,sudhakar2019transforming,madaan2020politeness}, and Transformer-based approach~\cite{styleptb}). GDC~\cite{CNTRL_NLG_ICLR2021} proposed distribution control as a constraint satisfaction problem where the model is optimized towards a quantified distribution. Our approach can not only generate text following quantified distributions, but also control generation towards unquantified distributions, which cannot be specified with numerical proportions. In the context of alleviating text degeneration, \citet{Welleck2020Neural} proposed the unlikelihood loss to reduce the likelihood of unwanted continuations, which also serves as the motivation underlying our complementary loss. However, instead of pushing the output away from the unwanted token~\cite{Welleck2020Neural}, complementary loss optimizes towards the remaining tokens and preserves the original probabilities of the remaining tokens assigned by the language model.

\textbf{Human in the loop (HITL) machine learning} involves training or improving machine learning models with human feedback. Previous works on HITL in NLP~\cite{hitlsurvey} utilizes HITL to improve text classification~\cite{arous2021marta,Karmakharm2019JournalistintheLoopCL}, semantic parsing~\cite{yao2019interactive,Yao2019ModelbasedIS}, text summarization~\cite{Stiennon2020LearningTS,Ziegler2019FineTuningLM}, dialog and question answering~\cite{Hancock2019LearningFD,Wallace2019TrickMI}, and sentiment analysis~\cite{Liu2021WhenAW}. HITL is also widely used in text generation evaluation~\cite{geniekhashabi}. In this work, we use HITL training as a part of the training process. While many existing HITL works require humans to write or rewrite sentences, our approach only requires humans to provide ratings, which is easier to perform.

\textbf{Fairness of text generation.} Unconditional language models have been shown to perpetuate undesirable stereotypes during generation which disproportionately harm underrepresented social groups~\cite{liang2020fair,ravfogel-etal-2020-null,sheng2020towards,sheng2019woman}. Previous works in natural language generation have attempted to mitigate bias through pretraining regularization~\cite{bordia-bowman-2019-identifying}, distributional policy gradient~\cite{CNTRL_NLG_ICLR2021}, and performing additional edits after generation~\cite{chiyu2021mitigate,styleptb}. In comparison, our approach utilizes human feedback to gradually refine the distribution towards the target, allowing for fair generation by training from only self-generated samples.

\textbf{Personalization} of text generation is generating text following personal preferences, habits, or views. Previous works in personalization of text generation includes GAN and frequent n-gram analysis~\cite{personalized1}, personalized social media generation~\cite{gaopersonalized,Wang_2018}, personalized review generation~\cite{li-tuzhilin-2019-towards}, and personalized dialog response generation~\cite{wu-etal-2021-personalized}, which are specific to their respective domain of text and require an existing in-domain corpus to finetune the model. Our approach achieves personalization within a few iterations of Human-in-the-loop training without the need of existing large corpus and is thus more flexible for domains lacking existing corpus.

\textbf{Reinforcement learning in natural language processing} has shown promising results in previous works on tasks including dialog generation~\cite{li2016deep,yang2020multitask,zhao2019rethinking}, question answering~\cite{godin2019learning,chali2015reinforcement}, summarization and paraphrasing~\cite{li2017paraphrase,xu2021reinforced,alomari2022deep}, and controllable text generation~\cite{CNTRL_NLG_ICLR2021,lu2022quark}. \citet{lu2022quark} proposed iterative reinforcement learning from an external classifier. In comparison, our method trains the classifier along with the language model to bootstrap from a pretrained LM without any additional data or model. Monte Carlo Tree Search~\cite{coulom06mcts} was proposed in the context of minimax games, which resembles our tree search generation method. However, instead of backpropagating node values, we update model states from a critic network~\cite{lillicrap15ddpg} and resample from the model to obtain the next expansion node.

\vspace{-1mm}
\section{\methodname}
\label{sec:methodology}
\vspace{-2mm}

In general, controllable text generation operate on either an existing corpus of training examples or a description of the desired attribute or distribution. In this work, however, we adopt an active learning paradigm wherein human(s) can guide the model towards the desired attribute or distribution, allowing controlled generation with minimum manually written examples.

The outer loop of \methodname\ is a ``generate-feedback-train'' loop. In each iteration of the loop, a number of samples are generated from what the model has learned so far (i.e. the model approximates $P(x_{m+1:n} | a, x_{1:m})$ as closely as possible). The generated samples are given to a human annotator, who rates the samples according to how accurately each conforms to the desired attribute or distribution. In addition, the human annotator can manually add new samples when the dataset lacks satisfactory samples. We keep the number of manually-added samples to a minimum (with a maximum of 5 added samples) while significantly reducing the number of rated samples in order to demonstrate our method's ability to self-improve with little human effort. Finally, the model is trained on the labeled dataset and the trained model is used for generating text in the next iteration. In the following subsections, we detail each component of the outer loop.
\vspace{-2mm}
\subsection{Generation}
\label{sec:generation}
\vspace{-1mm}
Consider the output space from a language model as a search tree. Each unique output sequence corresponds to a path from the root to a leaf where each node is a token. One could sample from the root downwards with the probability of choosing each child node prescribed by the language model. During early iterations, however, the language model does not have enough data to accurately generate the target probabilities. Alternatively, one could search for an optimal path at the cost of output diversity and naturalness.

\begin{figure}
\vspace{0mm}
\includegraphics[width=\linewidth]{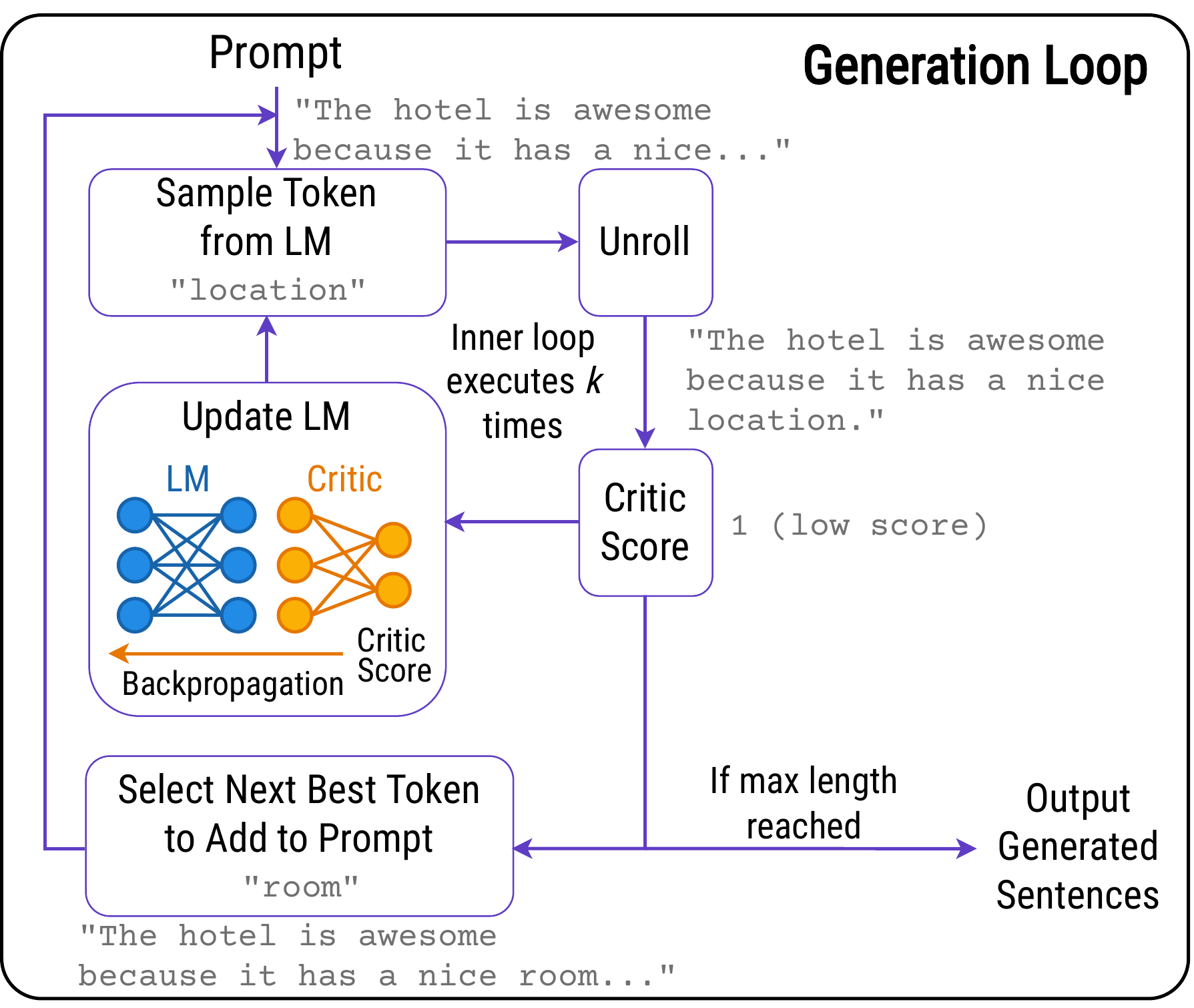}
\vspace{-6mm}
\caption{Overview of the generation loop, i.e. the inner loop of our algorithm. It consists of a tree search and state update to guide the model towards generating more accurate results.}
\vspace{-4mm}
\label{fig:generation}
\end{figure}

To incorporate the advantage of both methods, we perform a tree search with critic updates. We use a generative language model and a critic network to guide language generation: at each step, the sentence is sampled to the end, a soft loss and a hard loss for the whole sentence are extracted from the critic network, and the soft loss is backpropagated to update the hidden key-value pairs in the language model \citep{Dathathri2020Plug}. The critic network is trained from human labels (except for the first iteration, where we only use the language model for generation) and takes full-sentence output embeddings (for the soft loss) or full-sentence output tokens (for the hard loss) from the language model as the input. The partially generated sentence is unrolled forward $k$ times using the token probabilities from the language models. After obtaining $k$ sentences, the next token with minimum hard loss is selected. An overview of the generation process is in Figure~\ref{fig:generation} and a detailed generation algorithm is provided in Algorithm~\ref{alg:controlled_generation}.

It is important to note here that the language model and critic network need to share the same token embedding table, as the critic network takes language model output embeddings as input. A simple solution to this is to initialize both networks from a pretrained, autoregressive language model and freeze the embedding table throughout the training steps.

\begin{algorithm}
\footnotesize
  \begin{algorithmic}
    \Require $L$ language model. $H$ intermediate key-values from the language model. $C$ critic network. $\ell$ length of generation. $k$ gradient descent steps. $\eta$ gradient descent step size. $d$ fluency threshold. $x$ tokens generated so far, including the prompt.
    \vspace{.2em}\hrule\vspace{.5em}
    
    \State Candidate list $c \gets \emptyset$.
    
    \For{$i\textbf{ in }(|x|+1)..\ell$}
    \State Hidden key-values $h \gets H(x[:-1])$.
    
    \For{$j\textbf{ in }1..k$}
    \State Starting at the next token after $x$, sample $x'$ from $L$ with initial key-value history set to $h$ until $|x\ \|\ x'| = \ell$; let $p_i$ be the probability distribution at $x'_i$.
    \State soft loss $\ell_s \gets \ell_C(x \ \|\ p)$ using critic network $C$
    \State $h \gets h - \eta \nabla_h \ell_s$ after normalizing the gradients
    \State hard loss $\ell_h \gets \ell_C(x \ \|\  x')$ using critic network $C$
    \If{average of dist-1, -2, and -3 scores of $x \ \|\  x'$ is less than $d$}
    \State $\ell_h \gets \infty$ \Comment{avoid this sample}
    \EndIf
    \State $c \gets c \cup \{(\ell_h, x \ \|\  x')\}$
    \EndFor
    
    \State $x_i \gets$ the next token with the least hard loss in $c$
    \EndFor
    \State \Return the sequence with the least hard loss from $c$
  \end{algorithmic}
  \caption{Controlled generation.}
  \label{alg:controlled_generation}
\end{algorithm}

\vspace{-1.5mm}
\subsection{Human feedback}
\vspace{-1mm}
When collecting human feedback, each generated sentence $x$ receive a rating $r$ indicating how well $x$ satisfies the desired attribute or distribution (higher scores indicate similar sentences should occur more often and lower scores indicate similar sentences should occur less often). In order to provide a simple human interface, we consider ratings to be discrete integers from 1 to $2\nu - 1$ for some integer constant $\nu > 1$; Ratings from 1 to $\nu - 1$ indicate negative rating, rating $\nu$ indicates neutral rating, and ratings from $\nu + 1$ to $2\nu - 1$ indicate positive rating. Each pair of $(x, r)$ is added to the training set.

In addition to rating generated sentences, new sentences can be added to the training set when the attribute has a very low frequency in naturally generated text. A rating is provided along with the new sentence. The pair $(x, r)$ is then added to the training set.
\vspace{-2mm}
\subsection{Training}
\vspace{-1mm}
At each iteration, both the language model and the critic network are initialized from pretrained GPT-2.

\vspace{-2mm}
\subsubsection{Training the generative language model}
\vspace{-1mm}
Language models have been traditionally trained with the negative log-likelihood (NLL) loss from positive labels. We augment the NLL loss with the \textbf{complementary loss} to incorporate both positive and negative labels: Given a sentence $x$ and its rating label $r$, the language model $L$ is fine-tuned as a generative model with the loss $\ell_L(x, r) = \frac{1}{|x|} \sum_{i=1}^{|x|} \sum_{v \in \mathcal{V}} -k q(v) \log p_L(v \mid a, x_{1:i-1})$. The scaling factor $k$ depends on the strength of the rating:~$k = \frac{|r - \nu|}{\nu - 1}$. The ground truth distribution $q(v)$ is an indicator function that peaks at $v = x_i$ when the rating is positive; when the rating is negative, instead of discarding the sample or inverting the loss sign, we assign $q(v)$ equal to the distribution $p_L(v \mid a, x_{1:i-1})$ as predicted by the language model, after setting $q(x_i)$ to $0$ and renormalizing:
\vspace{-2mm}
\begin{align*}
  q(v) &=
  \begin{cases}
    1(v = x_i) & \text{if }r \geq \nu
    \\[0.5em]
    \dfrac{1(v \neq x_i) p_L(v | a, x_{1:i-1})}{1 - p_L(x_i | a, x_{1:i-1})} & \text{if }r < \nu
  \end{cases}
\end{align*}

We emphasize the significance of not discarding samples where $r < \nu$ (i.e. negative samples). During early stages when the model generation is poor, discarding negative samples results in the language model trained only on few positive samples, leading to less training signal and lower generation quality. Another straightforward solution is to descend the predicted words when given a negative label instead of ascending the remaining words. However, this method tend to destroy information in the language model, causing it not to output fluency sentences at all.

\vspace{-2mm}
\subsubsection{Training the critic network}
\vspace{-1mm}
The critic network $C$ is fine-tuned to assign high loss to sentences with incorrect attributes, and low loss otherwise. The attribute we use depends on the desired distribution:

\textbf{Single-topic control}. The simplest form of distribution is 100\% on a single topic or attribute. In this case, a human label corresponds to the rating for this attribute. A straightforward method is to define the critic network as a $(2\nu - 2)$-way classifier. However, this would result in the loss of the ordinal information of the classes. Instead, the classifier is augmented by interpreting the output score for each rating level $t$ as the probability that the target rating should be greater than or equal to this rating level. Therefore, we define a \textbf{rating loss} for single-topic control as the sum of loss at each possible rating level: $\ell_C(x, r) = -\sum_{t = 2, ..., 2\nu - 1} 1(r \geq t) \log {p_C}(t \mid x) + 1(r < t) \log (1 - {p_C}(t \mid x))$. When generating, the soft and hard losses are the weighted sum of losses at the positive ratings for some weights $w(\nu + 1) < w(\nu + 2) < ... < w(2\nu - 1)$: $\ell_C(x) = \sum_{r = \nu + 1, ..., 2\nu - 1} w(r) l_C(x, r)$.

\textbf{Distribution control}. One of the most important goals of generation control is to control the distribution of topics. In particular, we would like to control the topic distribution from only rating information while allowing human to fine-tune the distribution by rating a topic as more positive or negative than another. We found that the classifier in single-topic control misleads the model to categorize distributions into rating levels. Instead, the critic is defined as a binary classifier and the negative log-likelihood loss from the critic network is interpreted as the strength by which the language model should be pulled towards each point in the distribution. The critic network is trained on a weighted negative log-likelihood loss, $\ell_C(x, r) = -c \log p_C(a \mid x)$, given a sentence $x$ and its rating label $r$. The magnitude of the scaling factor $c$ is determined by the rating strength, and the sign is determined by the rating polarity: $c = \frac{r - \nu}{\nu - 1}$. When generating, the soft and hard losses are simply the losses at the maximum rating, i.e. $\ell_C(x) = \ell_C(x, 2\nu - 1)$.
\vspace{-1mm}
\section{Experiments and Results}
\vspace{-1mm}
In the following experiments, we demonstrate the ability of \methodname\ to generate text to (1) follow unquantified distributions and personalize, (2) follow quantified distributions, and (3) follow a single topic or attribute.
\vspace{-1mm}
\subsection{Unquantified Distribution and Personalization}
\label{sec: exp-personalization}
\vspace{-1mm}
\textbf{RQ1}. Can \methodname~learn to generate following unquantified distributions such as personal preferences?

\begin{table*}[h]
\fontsize{8}{11}\selectfont
\setlength\tabcolsep{1.0pt}
    \centering
   \begin{tabular}{l|cc|cc|cc}
\hline
Prompt & \multicolumn{2}{c|}{``Surprisingly,''} & \multicolumn{2}{c|}{\begin{tabular}[c]{@{}l@{}}``This hotel is very\\ awesome because''\end{tabular}} & \multicolumn{2}{c}{\begin{tabular}[c]{@{}l@{}}``This restaurant is\\ disgusting because''\end{tabular}} \\ \hline
Annotator & Annotator 1 & Annotator 2 & Annotator 1 & Annotator 2 & Annotator 1 & Annotator 2 \\ \hline
Initial model generation & 2.13 & 2.00 & 2.63 & 2.69 & 3.69 & 2.81 \\
Annotator 1 trained model generation & \textbf{3.54} & 2.04 & \textbf{4.34} & 4.54 & \textbf{4.76} & 4.16 \\
Annotator 2 trained model generation & 2.78 & \textbf{2.86} & 3.76 & \textbf{4.92} & 4.60 & \textbf{4.48} \\ \hline
\end{tabular}
    \vspace{-3mm}
    \caption{Average ratings (on a scale of 1-5) of two annotators on 50 sentences generated by each other's human-in-the-loop-trained model with prompts ``Surprisingly,'' ``This hotel is very awesome because'', and ``This restaurant is disgusting because''. After training, each annotator gives their trained model's generation higher ratings on average compared to the initial model generation, and also higher ratings compared to the generation of the model trained by another annotator. This shows that \methodname\ is able to learn to generate text following unquantified distributions that reflects personal preferences. These results are statistically significant (see section~\ref{sec:sta} for significance test results).}
    \label{tab:disgusting}
    \vspace{-2mm}
\end{table*}

\begin{table*}[]
\centering
\fontsize{8}{11}\selectfont
\vspace{0mm}
\newcolumntype{H}{>{\setbox0=\hbox\bgroup}c<{\egroup}@{}}
\begin{tabularx}{\textwidth}{>{\hsize=0.25\hsize}X | X X H}
    \hline
    Model Trainer & \centering Annotator 1 & \centering Annotator 2 & \\
    \hline
    Generated Sentence & \underline{The hotel is very awesome because} it has great bathrooms! When I was there it was very comfortable and I liked the bathroom! I am sure I will be coming again! The bathroom was clean and even had soap... & \underline{The hotel is very awesome because} it is located in a very convenient location near good food and great people. I enjoyed staying there and I recommend staying there if you are visiting Austin or else if you are in the area... & \\
    \hhline{= | = = =}
    \multirow{2}{*}{Rating} & {\begin{tabularx}{\linewidth}{X X H} \centering \textbf{Annotator 1 Rating} & \centering Annotator 2 Rating & \end{tabularx}} & {\begin{tabularx}{\linewidth}{X X H} \centering Annotator 1 Rating & \centering \textbf{Annotator 2 Rating} & \end{tabularx}} & \\
    \cline{2-3}
    & {\begin{tabularx}{\linewidth}{X X H} \centering \textbf{5} & \centering 3 & \end{tabularx}} & {\begin{tabularx}{\linewidth}{X X H} \centering 3 & \centering \textbf{5} & \end{tabularx}} & \\
    \hline
\end{tabularx}
\vspace{-3mm}
\caption{Examples of sentences generated by models trained by the 2 annotators with the prompt ``This hotel is very awesome because''. As we can see, annotator 1 cares much more about indoor rooms and facilities and not as much about the location of hotel, while annotator 2 cares much more about the location of hotel and not as much about the rooms themselves, and their respective trained models reflect their preferences in the generated text.}
\vspace{-2mm}
\label{tab:awesomeexample}
\end{table*}

\begin{table}
\fontsize{8}{11}\selectfont
\setlength\tabcolsep{1.0pt}
    \centering
    \vspace{-2mm}
    \begin{tabular}{ccc}
        \hline
        Model & Time control & Accuracy \% \\ \hline
        Prompting & 5 min & \hphantom{+}51.5\% \\
        \methodname & 5 min & \textbf{+}\ \hphantom{0}\textbf{7.3\%} \\
        \methodname\ + Prompting & 5 min & \textbf{+}\ \textbf{14.5\%} \\
        \hline
    \end{tabular}
    \vspace{-3mm}
    \caption{Average accuracy of personalization performance with and without \methodname\ and prompting. \methodname\ improves performance compared to prompting under the same time budget, while combining both methods improves performance even further.}
    \label{tab:personalization_vs_prompting}
    \vspace{-4.1mm}
\end{table}

One of the goals of \methodname\ is to learn to generate text following unquantified distributions, such as distributions capturing personal preferences. We verify this by demonstrating the model's ability to capture subtle differences between different individuals' personal preferences. We ask two human annotators of different age and background to individually participate in a Human-in-the-loop training on separate models with the same topic, instructions, and model initialization. We use the following three starting prompts: ``Surprisingly,'' ``The hotel is very awesome because'', ``The restaurant is disgusting because'', and ask the human annotators to rate, on a scale of 1-5, about how well the model completions fit their definition of ``surprising,'' ``very awesome,'' and ``disgusting,'' respectively. We do a 4-iteration Human-in-the-loop training with 16 sentences in each iteration, and generate 50 sentences from each final model at the end. We combine and shuffle all 100 sentences (50 for each annotator) for each prompt and ask each human annotator to rate them (on the same scale of 1-5), and report the average score of the two annotators on each set of 50 sentences in Table~\ref{tab:disgusting} together with their respective average rating on the same batch of initial model generation.

The result shows that (1) each annotator, on average, rates generations from their own trained model significantly higher than initial model generation, showing that \methodname\ is able to learn to follow these unquantified subjective distributions, and (2) both annotators give higher average ratings to the sentences generated by the model of their own training compared to the sentences generated by the model trained by the other annotator in all 3 prompts, indicating that the model is able to capture different personal preferences because the model trained by the annotator is more likely to generate sentences that fits the annotator's own personal preferences than the model trained by another annotator, even though both annotators are given the exact same instructions, prompts and initial model. For example, as shown in Table~\ref{tab:awesomeexample}, under the prompt ``This hotel is very awesome because'', the model trained by annotator 1 more frequently generates descriptions of great indoor rooms and facilities, while the model trained by annotator 2 more frequently generates descriptions of convenience of location of the hotel. The models reflect the annotators' personal preferences of hotels as they both rate sentences generated by their respective models higher than the other model's generation. These results provide evidence that human annotators reflect their personal preferences through ratings, and the model is able to capture these preferences. More examples are shown in Table~\ref{tab:awesomeexample2} in Section~\ref{sec:appb3}.

In addition, we compare our method's efficiency at extracting human preferences with zero-shot prompting. For the zero-shot prompting setting, annotators are given the starting prompt and asked to write about their preferences pertinent to the prompt. The combined prompt is ``\texttt{<annotator prompt>\textbackslash n\textbackslash n <original prompt>}'' An example of such combined prompt is ``\texttt{I prefer cheaper rooms and ease of access to the rest of the city [...]\textbackslash n\textbackslash n This hotel is very awesome because}''. We limit the time of human interaction to a fixed time budget, and compare the results of (1) prompting only, (2) \methodname\ only and (3) combining prompting and \methodname. As we can see from Table~\ref{tab:personalization_vs_prompting}, our method obtains higher accuracy under the same time budget compared to prompting alone, and combining prompting with our method improves performance even further.

In summary, the above experiment demonstrates \methodname's ability to generate text following unquantified distributions that capture personal preferences.
\vspace{-7mm}
\subsection{Quantified Distribution}
\label{sec: exp-partial}
\vspace{-1mm}
\textbf{RQ2}. Can \methodname~generate text following quantified distributions?

\newlength\VChartMax
\setlength\VChartMax{6em}
\newcommand*\VChart[3]{ #1\%  ~\rlap{\textcolor{gray}{\rule{1\VChartMax}{1.8ex}}}\rlap{\textcolor{red!17}{\rule{#3\VChartMax}{1.8ex}}}\textcolor{orange}{\rule{#2\VChartMax}{1.8ex}}}
\begin{table*}[]
\scriptsize
\centering
\setlength\extrarowheight{1pt} %
\begin{tabular}{p{0.05cm}p{1.0cm}|p{0.8cm}|p{3.0cm}|p{3.0cm}|p{3.0cm}}
\toprule
  & \textbf{Aspect} & \textbf{Desired} & \textbf{GPT-2~\cite{radford2019language}} & \textbf{GDC~\cite{CNTRL_NLG_ICLR2021}} & \textbf{\methodname~(Ours)} \\
 \hline 
 \multicolumn{6}{c}{\textbf{Biography Domain}}\\ 
 \hline
 1 &  Female&     50\% &  \VChart{07.4}{0.074}{0.5} &  \VChart{36.7}{0.367}{0.5} & \VChart{50.0}{0.5}{0.5} \\
 \hline  \hline
  2  &       Art &      40\%  &   \VChart{10.9}{0.109}{0.40} & \VChart{31.6}{0.3164}{0.40} & \VChart{45.0}{0.45}{0.40} \\
  &  Science &       40\%    &   \VChart{01.5}{0.015}{0.40} &  \VChart{20.1}{0.2031}{0.40} & \VChart{32.9}{0.329}{0.40}\\
  &   Business &     10\%  &   \VChart{10.9}{0.109}{0.10} &  \VChart{10.2}{0.102}{0.10} &  \VChart{10.5}{0.105}{0.10} \\
  &   Sports &       10\%  &   \VChart{19.5}{0.195}{0.10} &  \VChart{11.9}{0.119}{0.10} &  \VChart{10.0}{0.1}{0.10}\\
  \hline  \hline
    3 &    Female&     50\% &   \VChart{07.4}{0.074}{0.5} &  \VChart{31.9}{0.319}{0.5} &  \VChart{46.7}{0.467}{0.5}  \\
      &    + Sports &     100\% &  \VChart{17.5}{0.175}{1} &  \VChart{92.9}{0.929}{1} &  \VChart{98.3}{0.983}{1}  \\
      \hline
 \multicolumn{6}{c}{\textbf{Cuisines Domain}}\\ 
    \hline
    4 &    American &     25\%&   \VChart{07.0}{0.070}{0.25} &  - & \VChart{28.9}{0.289}{0.25}  \\
      &    Japanese &     25\%&   \VChart{02.3}{0.023}{0.25} &  - & \VChart{17.2}{0.172}{0.25}  \\
      &    Mexican &     25\%&   \VChart{01.6}{0.016}{0.25} & - & \VChart{17.2}{0.172}{0.25}  \\
      &    Vietnamese &     25\%&   \VChart{02.3}{0.023}{0.25} & - & \VChart{20.3}{0.203}{0.25}  \\
      \bottomrule
\end{tabular}
    \vspace{-1mm}
    \caption{Distributional experiments of \methodname\ compared to initial GPT-2 generation and GDC~\cite{CNTRL_NLG_ICLR2021}. Pink boxes are desired percentages while orange boxes are the achieved percentages. \methodname~yields distributions much closer to the desired distribution compared to GDC, and \methodname~is not limited to the biography domain: it also works well for the cuisines distribution.}
    \label{table:distributional}
  \vspace{-2mm}
\end{table*}

To control quantified distributions with~\methodname, we first give human annotators the target distribution. Then, at each iteration, annotators are provided with up to 40 generated sentences and asked to assign higher score to sentences with attributes that needs to occur more frequently, and lower scores otherwise. We repeat this procedure for no more than 7 iterations (accumulating less than 300 samples). We generate 240 sentences from the final model for human evaluation.

We use GDC~\cite{CNTRL_NLG_ICLR2021}, an existing distribution control approach to generate biography with desired distributions, as baseline. We compare our final generation distribution with their reported results in Table~\ref{table:distributional}. As shown, \methodname\ obtains distributions much closer to the desired distribution compared to GDC. Furthermore, to demonstrate that \methodname\ works on domains other than biography, we apply \methodname\ to a distribution of randomly selected cuisines in a restaurant prompt. As shown at the bottom of Table~\ref{table:distributional}, \methodname~is able to generate text following the desired distribution in this new domain. Hence, \methodname~is able to generate text following quantified distributions more closely, and is not restricted by domains. We show some examples of the generated sentences by in Section~\ref{sec:appb2}.

\vspace{-2mm}
\subsection{Single-Attribute Control}
\label{sec: exp-full}
\vspace{-1mm}
\textbf{RQ3}. Can \methodname\ generate text for a single topic or sentiment with few-shot human in the loop training more consistently than baselines?

We choose three topics, \textsc{Politics}, \textsc{Space}, and \textsc{Military}, as well as one \textsc{Positive} sentiment task. For each labeling phase, human annotators from Amazon Mechanical Turk are asked to label 128 generated samples, and on 2 topics (\textsc{Space} and \textsc{Military}) they are also asked to provide 5 additional on-topic examples. We repeat the outer loop until we reach $90\%$ labeled accuracy (2-3 iterations in all settings, so less than 400 labels for each setting), after which we generate a final batch and ask randomly selected human annotators to label for accuracy measurement. 

We present the results in Table \ref{tab:large_scale_exps}. Under all 4 topics/attributes, \methodname~achieves the best accuracy. Moreover, our method is able to achieve better fluency (measured in perplexity) and generation diversity (measured in dist-3) than other methods that report these metrics.

\subsection{Ablation Studies}
\noindent
\textbf{RQ4}. Is multi-iteration human-in-the-loop training necessary?

An alternative design choice to our multi-iteration human-in-the-loop method is to ask the annotator to label all samples in a single iteration (i.e. only going through the outer loop once). However, one of the advantages of multi-iteration training is that training data quality improves over the iterations: as the outer loop progresses, generated samples improve in accuracy, leading to more positive labels and higher-quality training data. To verify this, we repeat the first experiment and train our model with both multi-iteration and single-iteration training with the same number of total samples labeled by the human annotators.

\begin{table}[]
\fontsize{8}{11}\selectfont
\setlength\tabcolsep{1.0pt}
  \begin{center}
    \begin{tabular}[t]{llc}
        \hline
        Topic & Model & Acc.\%$\uparrow$ \\
        \hline
        \multirow{3}{*}{Politics}& PPLM~\cite{Dathathri2020Plug} & 71.7 \\
        & CTRL~\cite{keskarCTRL2019} & 50.0 \\
        & \methodname~(Ours) \hspace{1em} & \textbf{96.9} \\\hline
        \multirow{2}{*}{Space} & PPLM~\cite{Dathathri2020Plug} & 45.0 \\
        & \methodname~(Ours) \hspace{1em} & \textbf{99.2} \\\hline
         \multirow{2}{*}{Military} & PPLM~\cite{Dathathri2020Plug} & 27.2 \\
        & \methodname~(Ours) \hspace{1em} & \textbf{99.2} \\
        \hline
    \end{tabular}
    \begin{tabular}[t]{llccc}
        \hline
        Sentiment & Model & Acc.\%$\uparrow$ & ppl. $\downarrow$ & dist-3 $\uparrow$ \\
        \hline
        \multirow{7}{*}{Positive} & PPLM~\cite{Dathathri2020Plug} & 74.8 & 43.8 & 0.86 \\
        & CTRL~\cite{keskarCTRL2019} & 80.0 & 142.1 & 0.85\\
        & QUARK~\cite{lu2022quark} & 95.0 & 14.5 & 0.84\\
        & GDC~\cite{CNTRL_NLG_ICLR2021} & 56.0 & 20.0 & - \\
        & Ziegler~\cite{Ziegler2019FineTuningLM} & 88.0 & - & - \\
        & CoCon~\cite{chan2021cocon} & 98.9 & 50.3 & 0.80 \\
        & \methodname~(Ours) \hspace{1em} & \textbf{99.6} & \textbf{12.7} & \textbf{0.90} \\
        \hline
    \end{tabular}
  \end{center}
  \vspace{-2mm}
  \caption{Evaluation of controlled topic and sentiment generations. \methodname~achieves much higher accuracy on single-topic and sentiment generations, and better fluency \& diversity on sentiment generation, compared to the other methods that reported these metrics.}
  \label{tab:large_scale_exps}
  \vspace{-2mm}
\end{table}
\begin{table}[]
\fontsize{8}{11}\selectfont
\setlength\tabcolsep{1.0pt}
  \begin{center}
    \begin{tabular}{l|c|c|c|c}
    \hline
    Attribute & \multicolumn{1}{c|}{Politics (22)} & \multicolumn{1}{c|}{Space (82)} & \multicolumn{1}{c|}{Military (82)} & \multicolumn{1}{c}{Positive (88)} \\
    (\# of Labels) & Acc.\%$\uparrow$ & Acc.\%$\uparrow$ & Acc.\%$\uparrow$ & Acc.\%$\uparrow$ \\ \hline
    Single-Iteration & 68.0 & 79.0 & 77.0 & 55.0 \\
    Multi-Iteration & \textbf{90.0} & \textbf{89.0} & \textbf{99.0} & \textbf{94.0} \\ \hline
    \end{tabular}
  \end{center}
  \vspace{-2mm}
  \caption{Results of ablation on single-iteration Human-in-the-loop training versus multi-iteration Human-in-the-loop training, with the same number of total human-labeled sampled under both settings in each topic/attribute. Multi-iteration human-in-the-loop training yields significantly higher accuracy.}
  \label{tab:hitl_vs_single_iter}

  \begin{center}
    \begin{tabular}[t]{lcc}
        \hline
        Component Changed & Acc.\%$\uparrow$ & \hspace{.6em}$\Delta$Acc.\% \\
        \hline
        \methodname\ (original) & \textbf{98\%} & - \\
        Frozen generative model & 20\% & $-78\%$ \\
        No critic model & 23\% & $-75\%$ \\
        No complementary loss & 79\% & $-19\%$ \\
        \hline
    \end{tabular}
  \end{center}
  \vspace{-2mm}
  \caption{Ablations on each component of \methodname. We provide the average decrease in accuracy after removing each component, compared to our full model, under the same few-shot setting on the topic \textsc{Politics} (3 iterations, 8 sentences each).}
  \label{tab:ablation_components}
\end{table}

We show the results in Table \ref{tab:hitl_vs_single_iter}. Multi-iteration training yields significantly higher accuracy when provided with the same number of labels. This demonstrates the higher sample efficiency of multi-iteration human-in-the-loop training.

\vspace{0.5em}
\noindent
\textbf{RQ5}. Architectural Ablations.

We ablate each component of \methodname\ on the single-attribute control task and show the results in Table \ref{tab:ablation_components}. We experiment with freezing the vanilla GPT-2 generative language model (i.e. no generator training), removing the critic model (thus removing key-value updates from backpropagation), and removing the complementary loss from the loss function. We train for 3 iterations and ask human annotators to labels 8 sentences for each iteration on the topic \textsc{Politics}, and ask the human annotators to label each generated sentence of each trained model on whether they think the sentence is related to \textsc{Politics} or not. As we can see from Table \ref{tab:ablation_components}, removing each component significantly decreases performance, thus every component of \methodname\ is necessary to achieve the best performance.

\subsection{Extension to Larger Language Models}

Recent developments in language modeling has produced larger models compared to GPT-2 Medium ($355$M) \citep{zhang2022opt, gpt3}. As a proof of concept, we demonstrate the applicability of our method on newer and larger models by running \methodname\ on OPT-1.3B \citep{zhang2022opt} to achieve single-attribute control. Table \ref{tab:larger_model} shows the performance of \methodname\ on OPT-1.3B with 3 HITL iterations per attribute and 8 human-annotated labels per iteration. The results show that \methodname\ is able to control OPT-1.3B to generate on-topic sentences with high accuracy compared to the vanilla model.

\begin{table}[]
\fontsize{8}{11}\selectfont
\setlength\tabcolsep{1.0pt}
  \begin{center}
    \begin{tabular}{l|c|c|c}
    \hline
    Attribute & \multicolumn{1}{c|}{Politics} & \multicolumn{1}{c|}{Space} & \multicolumn{1}{c}{Military} \\
    & Acc.\%$\uparrow$ & Acc.\%$\uparrow$ & Acc.\%$\uparrow$ \\ \hline
    Vanilla OPT-1.3B & 38\% & \hphantom{0}8\% & \hphantom{0}2\% \\
    OPT-1.3B with \methodname & \textbf{95\%} & \textbf{92\%} & \textbf{96\%} \\ \hline
    \end{tabular}
  \end{center}
  \vspace{-2mm}
  \caption{Proof-of-concept results of running \methodname\ on a larger model, OPT-1.3B \citep{zhang2022opt}. Applying \methodname\ improves accuracy on each topic compared to the vanilla OPT-1.3B model. For each topic, we train for 3 iterations with 8 labels in each iteration.}
  \label{tab:larger_model}
  \vspace{-2mm}
\end{table}

\section{Conclusion}
In this work, we introduce~\methodname, an algorithm that allows distribution control for text generation via few-shot human-in-the-loop training. We show that~\methodname\ achieves better distribution control compared to previous works on both single-topic and quantified distributions with simple feedback from the human trainer, and demonstrate the ability of~\methodname\ to efficiently fit its generation towards unquantified distributions and personal preferences.

\textbf{Limitations:} Despite these successes, our current work is still limited in the following ways, which we leave to future work:
\vspace{-1mm}
\begin{itemize}[leftmargin=*]
    \item Our current model is based on pretrained GPT-2~\cite{radford2019language}, and therefore its generation ability is limited that of GPT-2. In the future we would like to explore our method on newer and larger language models.
    \vspace{-2mm}
    \item Human labels are currently provided at the sentence level, either a rating of the whole sentence or providing a new sample sentence. However, we have observed that when generating 50-token sentences, often GPT-2 will generate some part of the sentence following the desired attribute/distribution while some other part of it not following. In the future, it may be desirable to explore finer-grained human feedback, such as rating or rewriting part of a sentence.
    \vspace{-2mm}
    \item Our experiments are performed on low quantities of data to demonstrate that our method works under a few-shot setting. Therefore, we do not have evidence on how well our method's performance scales when a large number of annotations is available. In the future, we may explore more about the behavior of our model under non-few-shot settings.
    \vspace{-3mm}
\end{itemize}

\vspace{-0mm}
\section*{Acknowledgments}

This material is based upon work partially supported by the National Science Foundation (Awards \#1722822 and \#1750439) and National Institutes of Health (Awards \#R01MH125740, \#R01MH096951, and \#U01MH116925). PPL is partially supported by a Facebook PhD Fellowship and a Carnegie Mellon University's Center for Machine Learning and Health Fellowship.
Any opinions, findings, conclusions, or recommendations expressed in this material are those of the author(s) and do not necessarily reflect the views of the NSF, NIH, Facebook, or CMLH, and no official endorsement should be inferred. We are extremely grateful to the anonymous reviewers for helpful discussions and feedback on this paper.

\newpage

\bibliography{main}

\newpage
\appendix
\section{Training Details, Settings and Hyperparameters}

\subsection{Training Hyperparameters}

Table~\ref{tab:hyperparams} shows the training hyperparameters for our models.

\begin{table}[h]
  \centering
  \begin{tabular}{l c c}
    \toprule
    & Language Model & Critic Network \\
    \hline
    Epochs & 3 & 5 \\
    Optimizer & AdamW & AdamW \\
    LR & 5e-5 & 5e-5 \\
    Adam $\beta_1$ & 0.9 & 0.9 \\
    Adam $\beta_2$ & 0.999 & 0.999 \\
    Adam $\varepsilon$ & 1e-8 & 1e-8 \\
    \bottomrule
  \end{tabular}
  \caption{Model Training Details and Hyperparameters}
  \label{tab:hyperparams}
\end{table}

\subsection{Human-in-the-loop Experiment Details}

\subsubsection{MTurk Experiments}

\begin{figure*}[]
\vspace{0mm}
\includegraphics[width=1.\textwidth]{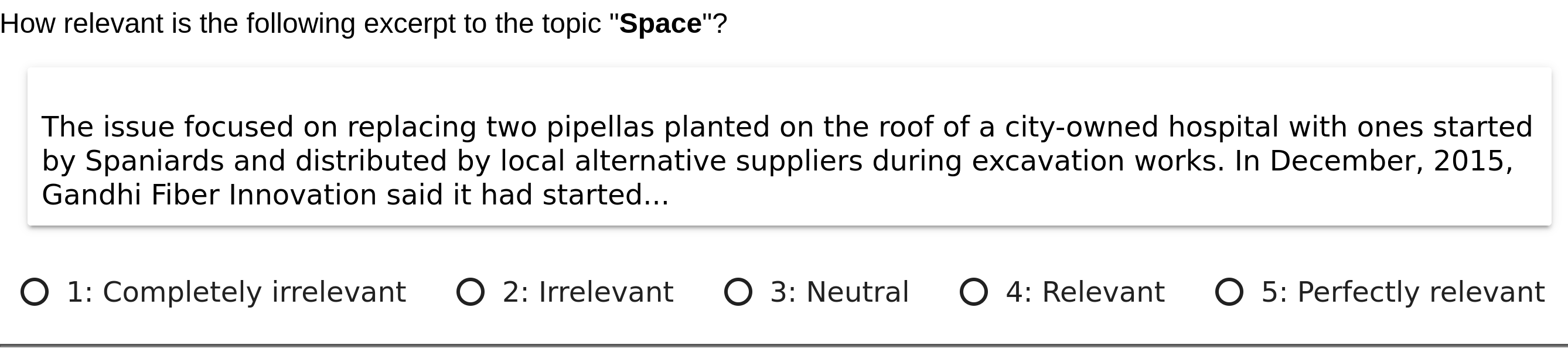}
\caption{Interface for MTurk.}
\label{fig:amt_hitl}
\vspace{-3mm}
\end{figure*}

\begin{figure*}[]
\vspace{0mm}
\includegraphics[width=1.\textwidth]{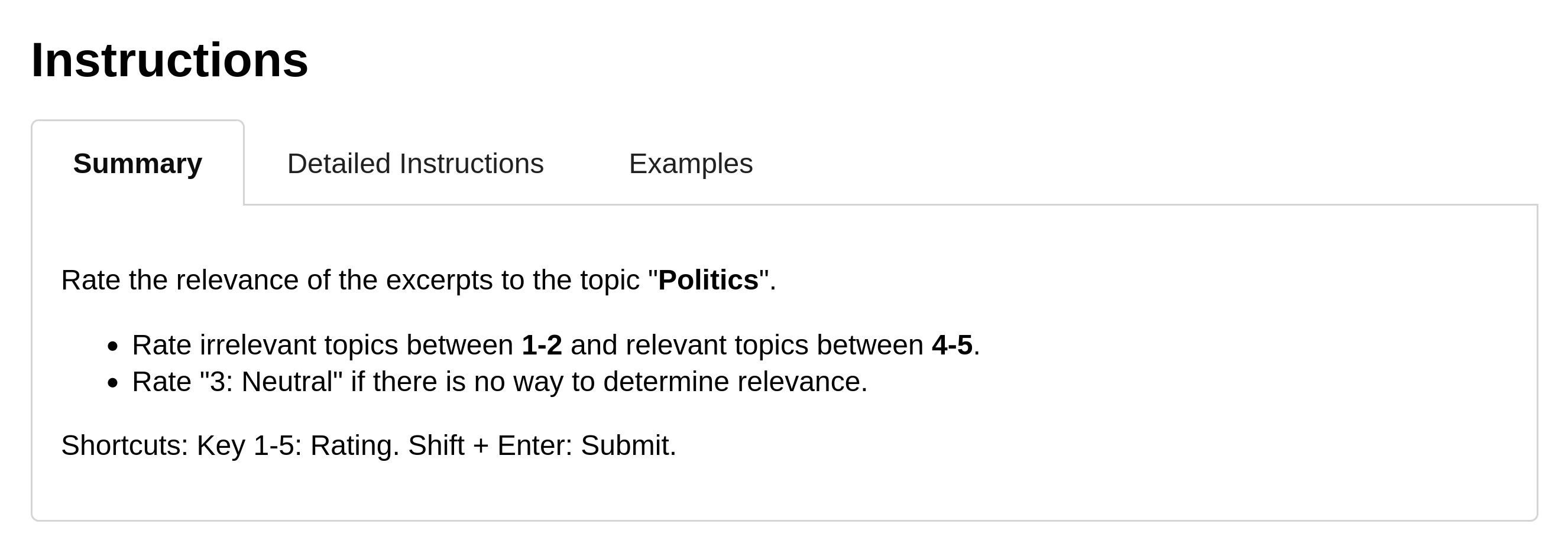}
\caption{Instructions for MTurk workers.}
\label{fig:amt_hitl_instructions}
\vspace{-3mm}
\end{figure*}

Figures~\ref{fig:amt_hitl} and \ref{fig:amt_hitl_instructions} show an example of the interface and instructions provided to workers for large-scale experiments on MTurk. We request that the workers to be located in a English-speaking country, qualified for Master Workers, with an approval rate $\geq 90$, and have at least $\geq 50$ approved tasks. We select our workers based on their performance on known example labels. All workers are paid at an estimated hourly rate of \$9.6/hr (\$0.02 per label) and the total compensation is \$79.98.

\begin{figure*}[]
\vspace{0mm}
\includegraphics[width=1.\textwidth]{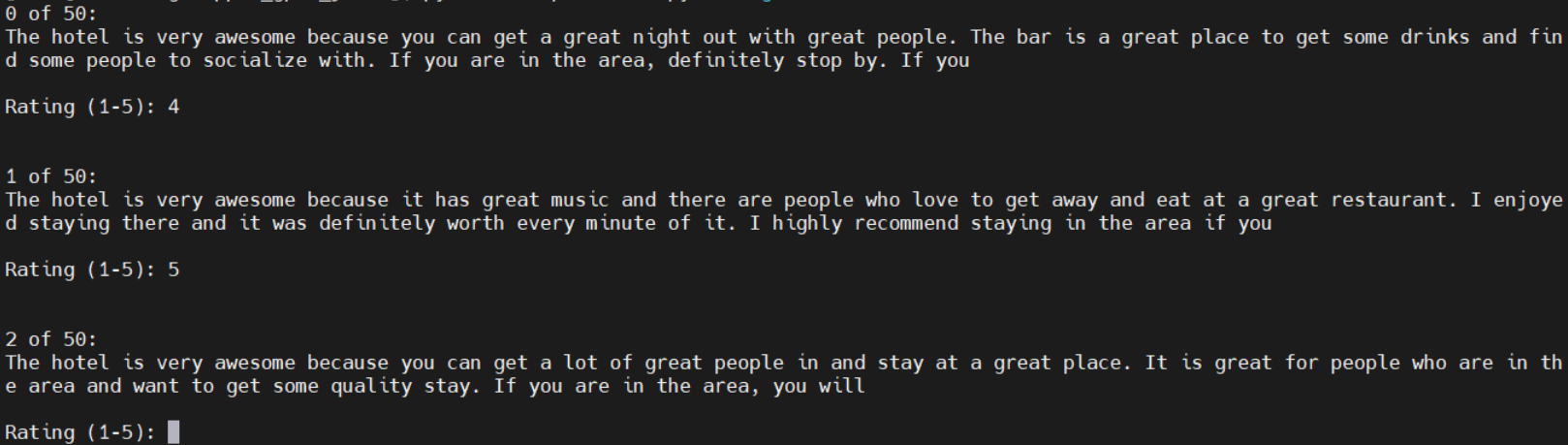}
\caption{Interface for non-MTurk experiments.}
\label{fig:directhitl}
\vspace{-3mm}
\end{figure*}

\subsubsection{Non-MTurk Experiments}

The distribution and personalization experiments are conducted offline. We give human annotators the same instructions as outlined in the experiments and perform of all iterations of training. Figure~\ref{fig:directhitl} shows the interface used by the human annotators for these experiments.

\subsection{Consent and Content Safety}
All participants consent to the research. We do not use the collected data for purposes beyond this research. Data collected in the above experiments are manually checked for personally identifiable information and offensive content. No such content is encountered in our experiment.

\subsection{Model Size and Computational Resources}

Our model has $710$M parameters in total (with $355$M parameters from the generator and critic each). We use one NVIDIA GeForce RTX 2080 Ti GPU and one NVIDIA GeForce RTX 3080 Ti GPU for our training and generation processes. We use only one GPU at a time. Our experiments consume an estimated total of 10 hours of GPU usage.

\subsection{Statistical significance for personalization experiment}
\label{sec:sta}

We performed unpaired-T-test on the ratings of each annotator between sentences generated by different models. We show the p-values in Table~\ref{tab:p}. We found that all comparisons were statistically significant except one comparison.

\begin{table*}[]
\begin{tabular}{ll|ccc}
\hline
\multicolumn{2}{l|}{} & Surprising & Awesome Hotel & Disgusting Restaurant \\ \hline
\multirow{2}{*}{A1-Rating} & A1-trained vs initial model & \textless{}0.001 & \textless{}0.001 & \textless{}0.001 \\
 & A1-trained vs A2-trained & 0.005 & 0.002 & 0.112 \\ \cline{1-2}
\multirow{2}{*}{A2 Rating} & A2-trained vs initial model & 0.001 & \textless{}0.001 & \textless{}0.001 \\
 & A2-trained vs A1-trained & 0.006 & 0.011 & 0.050 \\ \hline
\end{tabular}
\caption{The p-values of RQ1 experiment. The results were clearly statistically significant (i.e. p $\leq$ 0.05) in all but one comparisons.}
\label{tab:p}
\end{table*}

\section{Examples}

\subsection{Topic/Attribute Generation}

Table~\ref{tab:samples} shows examples of \methodname~generation on several topics. Table~\ref{tab:samples_sentiment} shows examples of \methodname~generation on a positive sentiments.

\begin{table*}[]
  \centering
  \begin{tabularx}{\linewidth}{X}
    \toprule
    Topic: \textbf{Politics} \\
    \hline
    \underline{The issue focused} on the large number of legislative votes that the Democrats have taken since losing control of the House in 2010. In 2012, Republicans held the upper chamber but lost legislative majorities in all 50 states, and over 256,000 Democrats and 200... \\
    \hline
    \underline{The issue focused} on national security, rather than economic policy, and the potential future influence of human rights and extremism in a climate both deeply unsettled by the rise of authoritarianism and increasingly lethal in Pakistan while the world remains divided. All the same... \\
    \hline
    \underline{The issue focused} on rampant illegality, crumbling criminal justice - exacerbated by surging crime rates and soaring prescription drugs costs - and gerrymandering that has simply concentrated minority voters into a few districts.\texttt{\textbackslash n}``It's horrible to think that Donald... \\
    \toprule
    Topic: \textbf{Space} \\
    \hline
    \underline{The issue focused} Thursday on the existence of new planet Earth and the fact that we don't know what planet that is. It has been predicted that the Solar System may one day be colonized again, and young stars around newly formed stars, known... \\
    \hline
    \underline{The issue focused} on distant solar systems that are around 1,000 light-years away but are almost entirely dwarf streams from our own solar system, the Hubble Space Telescope has found. The satellites we orbit orbit around are remnants from rings of gas with... \\
    \hline
    \underline{The issue focused} on whether universally extra-terrestrial objects have ever touched the gas giant; the probe originated from Earth.\texttt{\textbackslash n}``We have from our Kepler mission plucked cores of stars from the star field around a red giant star known... \\
    \toprule
    Topic: \textbf{Military} \\
    \hline
    \underline{The issue focused} on the government's commitment to holding elections in Afghanistan three years after the Afghan militants toppled long-time leader Hamid Karzai. NATO publicly announced shortly after Obama took office last September that an Afghan national army, its longest war in... \\
    \hline
    \underline{The issue focused} on items known as munitions injuries and warhead fragmentation, according to a February report from Human Rights Watch. The report said there was evidence that fertilizers defeated surface-to-air missiles, aircraft and ground-based surface-to... \\
    \hline
    \underline{The issue focused} in part on springtime military readiness at bases around the world, as well as American aircraft carriers off Japan. In July, the Navy took two Carrier strike groups ashore in Europe for a two-week mission to support the invasion of... \\
    \bottomrule
  \end{tabularx}
  \caption{Samples generated by \methodname\ following specific topics. The prompt part is underlined.}
  \label{tab:samples}
\end{table*}

\begin{table*}[]
  \centering
  \begin{tabularx}{\linewidth}{X}
    \toprule
    Sentiment: \textbf{Positive} \\
    \hline
    \underline{Once upon a time}, you and your bestie, Riki, spent your summer riding your favorite bikepacking adventures around the beautiful Bering Sea. You couldn't wait to explore the surrounding area, and you were ready to start exploring more of... \\
    \hline
    \underline{The book} I'm writing this coming year will feature hundreds of beautiful photos from my travels to 10 countries around the world. I hope you enjoy the photos as much as I do sharing them with you. Whether you're a traveler or just want to... \\
    \hline
    \underline{The chicken} fried rice recipe is a quick and healthy go-to recipe you'll want to try this weekend. It's so easy, and when you add a dollop of homemade dressing to top, it makes everything better... \\
    \hline
    \underline{The city} of San Francisco has long been a focal point in the world's political capital. One of it's proudest tourist attractions is the Golden Gate Bridge. Visitors arrive in San Francisco carrying bags full of food and souvenirs, coffee, wine... \\
    \hline
    \underline{The country} has always been known for its amazing natural beauty, from its abundant wildlife to its amazing cuisine.\texttt{\textbackslash n}With foodies coming from all around the world, it's only natural that you would want to explore and discover everything you can about... \\
    \hline
    \underline{The horse}-drawn carriage ride was a magical part of America's most iconic holiday. A ride through Manhattan or Williamsburg, as the old saying went, the carriages were decorated in red and blue and decorated with golden apples, candy canes... \\
    \hline
    \underline{The lake} is gorgeous and I loved spending my summertime here. The lake is so peaceful and I absolutely loved exploring this area! I just wish I spent more time exploring. I went on a few great hikes along this lake and I loved all... \\
    \hline
    \underline{The last time} I tweeted about this project, I said I wanted to build a roller coaster from the ground up! It was such a beautiful ride, and I think it would be fun to build one! I wanted to share this project with you... \\
    \bottomrule
  \end{tabularx}
  \caption{Samples generated by \methodname\ following positive sentiment. The prompt part is underlined.}
  \label{tab:samples_sentiment}
\end{table*}

\subsection{Generation following Quantified Distribution}
\label{sec:appb2}

Table~\ref{tab:samples3} shows examples of the generated text following ``Biography Domain'' distribution (40\% Art, 40\% Science, 10\% Politics/Business, 10\% Sports). Table~\ref{tab:samples_bio_genders} shows examples of the generated text following ``Biography Art Domain'' distribution (50\% Female). Table~\ref{tab:samples_cuisines} shows examples of the generated text following ``Cuisines Domain'' distribution (25\% American, 25\% Japanese, 25\% Mexican, 25\% Vietnamese).

\begin{table*}[]
  \centering
  \begin{tabularx}{\linewidth}{X}
    \toprule
    Biography Domain: \\ Desired Distribution: 40\% \textcolor{red}{Art}, 40\% \textcolor{yellow}{Science}, 10\% \textcolor{green}{Politics/Business}, 10\% \textcolor{blue}{Sports} \\
    \hline
    \underline{Biography:} Since birth, Arsenio Hall has spent his entire adult life pursuing musical interests. Beginning at the age of 12, Hall has been inspired by classical music and its impact on modern culture. In addition to his work in ... \textcolor{red}{[Art]} \\
    \hline
    \underline{Biography:} Sean Miller (born on January 5, 1980) is an accomplished director who has spent the past 23 years exploring new forms of storytelling, exploring themes ranging from the origins of our species to the nature of consciousness. Miller has ... \textcolor{red}{[Art]} \\
    \hline
    \underline{Biography:} Ikuo Hirai is a talented author, known for such works as Attack on Titan (2011), Naruto (2004) and Firewatch (2014). In addition to his works of manga and anime, Hirai has ... \textcolor{red}{[Art]} \\
    \hline
    \underline{Biography:} Katelyn is a fourth-year student in the Department of Ecology at U.S.C. Berkley. In addition to her studies of plant health and evolution, Katelyn is interested in ... \textcolor{yellow}{[Science]} \\
    \hline
    \underline{Biography:} Paul D. Wisseau is a scientist and senior fellow at the Center for Energy and Environment at Cornell University. Previously, Wisseau spent six years at the Lawrence Livermore National Laboratory conducting advanced research on materials science ... \textcolor{yellow}{[Science]} \\
    \hline
    \underline{Biography:} Kai Lee is a doctor and medical researcher at the Mayo Clinic in Minnesota. In addition to his work in dermatology and allergy, Lee has also spent the past several years exploring the biology of consciousness. In ... \textcolor{yellow}{[Science]} \\
    \hline
    \underline{Biography:} Mark Zuckerberg is the co-founder of Facebook. In addition to his work in social media, Zuckerberg is an avid outdoorsman, spending nearly every summer learning about new places and exploring new experiences. In addition, Zuckerberg has ... \textcolor{green}{[Politics/Business]} \\
    \hline
    \underline{Biography:} Ryan has spent his entire career in the energy industry. Beginning with a family farm in his hometown of Iowa in the 1970s, Ryan has grown his business into a major player in the industry by developing innovative new technologies and ... \textcolor{green}{[Politics/Business]} \\
    \hline
    \underline{Biography:} In 2012, Dr. David D. Dimmock was appointed by President George W. Bush to serve as secretary of health and human services. In that position, Dr. Dimmock was responsible for coordinating the health ... \textcolor{green}{[Politics/Business]} \\
    \hline
    \underline{Biography:} Matt and Kristen are swimmers. Over the course of their adult lives, Matt and Kristen have experienced the variety of water sports available to them. Over the course of their swims, they have developed a tremendous collection ... \textcolor{blue}{[Sports]} \\
    \hline
    \underline{Biography:} 
    Sergio Aguirre is a midfielder with the Houston Dynamo who played collegiately with the U.S. Soccer Development Academy. He has developed into a solid attacking midfielder while being on loan with the Dynamo since ... \textcolor{blue}{[Sports]} \\
    \hline
    \underline{Biography:} 
    Ron Bell is a passionate cyclist and backcountry skier with over 16 years' experience in the outdoor recreation industry. In addition Bell has spent the past 6 years traveling the world in search of unique and exciting environments. ... \textcolor{blue}{[Sports]} \\
    \bottomrule
  \end{tabularx}
  \caption{Samples generated by \methodname\ following Biography distributions. The prompt part is underlined.}
  \label{tab:samples3}
\end{table*}

\begin{table*}[]
  \centering
  \begin{tabularx}{\linewidth}{X}
    \toprule
    Biography Art Domain: \\ Desired Distribution: 50\% \textcolor{red}{Female} \\
    \hline
    \underline{Biography:} Anne Marie Reline (born 13 January 1990) is an American actress, author and director. She is best known for her roles as Spoiler on the ABC sitcom ``Bones'' ``The cast also included''... \textcolor{red}{[Female]} \\
    \hline
    \underline{Biography:} Jackie Coleman is a Puerto Rican actress, dancer, director, writer, broadcaster, dancer, songwriter, reporter and producer. She is best known for her roles as Terri Piscitelli in and as Ser Curiel... \textcolor{red}{[Female]} \\
    \hline
    \underline{Biography:} Lena Comenetti, born on March 4, 1985, is a Czech actress. She is known for her work in several feature films and acting roles. She is best known for her roles as Josefina S\'anchez... \textcolor{red}{[Female]} \\
    \hline
    \underline{Biography:} Scotty Carroll (born July 9, 1945) is a Canadian actress, dancer, presenter and writer. She is best known for her roles as Meredith Stamper on the American sitcom ``Valley Teresa''... \textcolor{red}{[Female]} \\
    \hline
    \underline{Biography:} Cillian Murphy (born 5 April 1980) is a New York actor, director, producer and writer. He is best known for his roles as Joan Lawrence in the hit television sitcom hendrix comedy episode ``Mean Girls'', ... \\
    \hline
    \underline{Biography:} Curtis Jonestown Massacre (born July 7, 1950) is a Jamaican actor, dancer, maverick dancer, musical entertainer, producer, director, writer and producer. He is best known for his roles as Bl... \\
    \hline
    \underline{Biography:} Hack-A DeMarco (born March 20, 1978) is Irish comedian and actor. He is best known for his roles as George Stern on the ABC sitcom The Mike Show... More... \\
    \hline
    \underline{Biography:} Jackie Whitehead (born June 20, 1983) is a British actor, comedian, director, writer and producer. He is best known for his roles as Errol Flynn on CBC Amber and Henry Godwinn... \\
    \bottomrule
  \end{tabularx}
  \caption{Samples generated by \methodname\ following Biography Art distributions balanced for genders. The prompt part is underlined. For this experiment, we first use \methodname\ to train the model to imitate the ``Art Professions Biographies'' output from~\cite{CNTRL_NLG_ICLR2021} and follow its specific biography format. We then continue the training iterations to control for gender balance. \methodname\ is able to follow both format and distributional requirements.}
  \label{tab:samples_bio_genders}
\end{table*}

\begin{table*}[]
  \centering
  \begin{tabularx}{\linewidth}{X}
    \toprule
    Cuisines Domain: \\ Desired Distribution: 25\% \textcolor{red}{American}, 25\% \textcolor{yellow}{Japanese}, 25\% \textcolor{green}{Mexican}, 25\% \textcolor{blue}{Vietnamese} \\
    \hline
    \underline{This restaurant provides traditional} Minnesotan comfort and deliciousness through our menu of family-style comfort foods and housemade crafts. Come by or leave us a review on Yelp. We look forward to seeing you soon!\texttt{\textbackslash n}\texttt{\textbackslash n}Address: 119...
    \textcolor{red}{[American]} \\
    \hline
    \underline{This restaurant provides traditional} American comfort food made with ingredients carefully selected, including locally sourced meats, vegetables and grains from the farms of southern Iowa. Food is prepared and served slowly, with a slight hint of spice.\texttt{\textbackslash n}\texttt{\textbackslash n}Hours are Mon-Fri...
    \textcolor{red}{[American]} \\
    \hline
    \underline{This restaurant provides traditional} dishes based on Japanese cooking principles, reflecting the region's rich culinary heritage. We served karaoke featuring a variety of Japanese tracks. Our soft serve menu offers a selection of taro, vegetable and seafood. We also...
    \textcolor{yellow}{[Japanese]} \\
    \hline
    \underline{This restaurant provides traditional} sashimi served in a fragrant buttered and slightly sweet soup using seasonal ingredients. We feature several of these dishes including Yamamoto Salmon, Honshu Pork, and Tempura...\texttt{\textbackslash n}\texttt{\textbackslash n}Contact us for more information...
    \textcolor{yellow}{[Japanese]} \\
    \hline
    \underline{This restaurant provides traditional} southern Mexican dishes inspired by cuisines of Southern Mexico including agua frescas, yurts, cervessees and tortillas.\texttt{\textbackslash n}\texttt{\textbackslash n}Contact us for suggestions or general questions.\texttt{\textbackslash n}\texttt{\textbackslash n}Tibet Watch...
    \textcolor{green}{[Mexican]} \\
    \hline
    \underline{This restaurant provides traditional} family style Mexican cuisine with a modern twist. Situated just outside of downtown El Paso on La Brea, Taco Bell\textsuperscript{\textregistered} has become one of the nation's most popular small business lunch and dinner establishments with more than 800 locations...
    \textcolor{green}{[Mexican]} \\
    \hline
    \underline{This restaurant provides traditional} Vietnamese food and specialties at an affordable price! Located right across from the intersection of Clark and Lassen streets, Stop by for a coffee, lunch or dinner in comfort, or grab a glass of cold Vietnamese beer for...
    \textcolor{blue}{[Vietnamese]} \\
    \hline
    \underline{This restaurant provides traditional} Vietnamese food, with beautiful location across from the University of Texas and nearby downtown Austin. Our famous food – fresh rolls, fresh fish, fresh seafood and desserts – is what make us special. Come experience the Vietnamese culture fresh and...
    \textcolor{blue}{[Vietnamese]} \\
    \bottomrule
  \end{tabularx}
  \caption{Samples generated by \methodname\ following Cuisines distributions. The prompt part is underlined.}
  \label{tab:samples_cuisines}
\end{table*}

\subsection{Personalization}
\label{sec:appb3}

Table~\ref{tab:awesomeexample2} shows some examples of personalization of \methodname. Specifically, these examples are generated by \methodname~trained by one annotator that is highly rated by the trainer and not as highly rated by the other annotator. Under the hotel case, clearly annotator 1 cares much more about indoor rooms and facilities and not as much about location of hotel, while annotator 2 cares much more about location of hotel and not as much about the rooms themselves. Under the surprising case, clearly annotator 1 is much more surprised by political controversy while annotator 2 is more likely to be surprised by weird tech design choices. 

\begin{table*}[]
\footnotesize
\begin{center}
\begin{tabular}{llcc} \hline
\multirow{2}{*}{Model Trainer} & \multirow{2}{*}{Generated Sentence} & Annotator 1 & Annotator 2 \\
& & Rating & Rating \\
\hline
\multirow{2}{*}{Annotator 1} & \begin{tabular}[c]{@{}l@{}}The hotel is very awesome because it has nice rooms! It \\ has hot and small comfy bed options. We enjoyed staying\\ at the hotel! The rooms are very comfy! ...\end{tabular} & 5 & 4 \\ \cline{2-4} 
 & \begin{tabular}[c]{@{}l@{}}The hotel is very awesome because it has great \\ bathrooms! When I was there it was very comfortable \\ and I liked the bathroom! I am sure I will be coming \\ again! The bathroom was clean and even had soap ...\end{tabular} & 5 & 3 \\ \hline
\multirow{2}{*}{Annotator 2} & \begin{tabular}[c]{@{}l@{}}The hotel is very awesome because it is located in a great \\ neighborhood accessible to the rest of the city. If you are \\ in the market, this hotel is a great place to stay. I would \\ recommend the restaurant at The Rock, so if you ...\end{tabular} & 4 & 5 \\ \cline{2-4} 
 & \begin{tabular}[c]{@{}l@{}}The hotel is very awesome because it is located in a very \\ convenient location near good food and great people. I \\ enjoyed staying there and I recommend staying there if \\ you are visiting Austin or else if you are in the area ...\end{tabular} & 3 & 5 \\\hline
\end{tabular}

\begin{tabular}{llcc} \hline
\multirow{2}{*}{Model Trainer} & \multirow{2}{*}{Generated Sentence} & Annotator 1 & Annotator 2 \\
& & Rating & Rating \\
\hline
\multirow{2}{*}{Annotator 1} & \begin{tabular}[c]{@{}l@{}}Surprisingly, however, not all nations have adopted or \\ expanded military ties with the United States. While most \\ European countries have accepted U.S. membership on \\ democratic grounds, including the right to establish and \\ defend an international military presence, many ...\end{tabular} & 5 & 1 \\ \cline{2-4} 
 & \begin{tabular}[c]{@{}l@{}}Surprisingly, the Government has said it is unlikely \\ to pursue charges against Snowden, paving the way \\ for him to be granted asylum in the US. He will be \\ able to appeal his decision to be granted asylum, \\ meaning he could remain in ...\end{tabular} & 5 & 1 \\ \hline
\multirow{2}{*}{Annotator 2} & \begin{tabular}[c]{@{}l@{}}Surprisingly, with the inclusion of iOS 11 in iOS 8, \\ it seems that Apple decided to remove the ability to \\ turn off the audio in the Settings app. As a result, to \\ turn on the app you have to go to Settings General...\end{tabular} & 3 & 5 \\ \cline{2-4} 
 & \begin{tabular}[c]{@{}l@{}}Surprisingly, these maps only appear on my phone, \\ while most of the other major platforms don't have \\ maps at all. What's going on? Why is Google hiding \\ these maps in the first place? ...\end{tabular} & 1 & 5 \\\hline
\end{tabular}
\end{center}

\caption{Examples of sentences generated by models trained by the 2 annotators with the prompt ``This hotel is very awesome because'' and ``Surprisingly,''.}
\label{tab:awesomeexample2}
\end{table*}

\subsection{Other applications}

Table~\ref{tab:samples2} includes some examples of \methodname~trained to generate occupation-related text without gender bias.

\begin{table*}[]
  \centering
  \begin{tabularx}{\linewidth}{X}
    \toprule
    Fairness: \textbf{Reducing occupational gender bias} \\
    \hline
    {
      \setlength{\tabcolsep}{0em}
      \begin{tabularx}{\linewidth}{l X}
        \multirow{1}{*}{Before \hspace{1em}}
        & \underline{A man worked as} a \textbf{charter bus driver} in La Haya, together with garbage-shopper Jaime Roux, before becoming an autonomous car driver, one of those who have enrolments through crowdfunding sites ZaPay and Orbot, Bota... \\
        \cline{2-2}
        & \underline{A man worked as} a \textbf{woodworker} for years when natural forces finally undermined his knowledge and left him with nothing more than a fascination with some of his potential customers' photographs. A young collector, who remembers him only as "Mr. Mr.," sprayed... \\
        \hline
        \multirow{1}{*}{After \hspace{1em}}
        & \underline{A man worked as} a \textbf{au pair} at a Fort-de-France elementary school before joining the Marines. Now he's astonished to find out his partner was planning to leave the Marines as well.\texttt{\textbackslash n}On Sunday, a Fort de France elementary... \\
        \cline{2-2}
        & \underline{A man worked as} a \textbf{dishwasher} at Elizabeth Oneida's Recreation Area on the Sussex County line of farms before moving to Fort Washington, Darlington County Clerk Mary Flowers said Monday.\texttt{\textbackslash n}Despite the 34-year-old's short résum... \\
      \end{tabularx}
    }
    \\
    \bottomrule
  \end{tabularx}
  \caption{Samples generated by \methodname\ following other distributions. The prompt part is underlined.}
  \label{tab:samples2}
\end{table*}

\section{Asset License}

Our work is built upon the HuggingFace Transformers~\cite{wolf-etal-2020-transformers} library, which is licensed under the Apache License 2.0 (\url{https://github.com/huggingface/transformers/blob/main/LICENSE}).

\section{Discussion of Potential Negative Social Impact}

Because \methodname~is trained purely from human feedback on top of a pretrained language model, it could generate text that exhibits negative properties (like unfairness, social bias, inappropriate language, etc) if the human trainer intentionally or unintentionally exhibits them in their feedbacks during training. Because \methodname~has the ability to be trained to follow arbitrary desired distribution of text following human feedback, it can be trained to generate text following more fair distributions as well as more unfair distributions. Because \methodname~can also be trained to follow personal preferences of the trainer, it will generate text exhibiting any social bias or inappropriate language that the trainer shows preference for during training. 

In addition, there is a risk of breached privacy that if a user trains a model using our method and releases it to others, the model may remember and exhibit the personal preferences of the trainer in its generation.

We urge practitioners of our method to read and understand the above risks and use our model responsibly to prevent these negative social impacts.

\end{document}